\documentclass{article}

\usepackage[utf8]{inputenc}
\usepackage[T1]{fontenc}
\usepackage{lmodern}
\usepackage[margin=1in]{geometry}
\usepackage{amsmath,amssymb}
\usepackage{booktabs}
\usepackage{graphicx}
\usepackage{hyperref}
\usepackage{caption}
\usepackage{float}

\usepackage{titlesec}
\titleformat{\section}{\bfseries\large}{}{0em}{}
\titleformat{\subsection}{\bfseries}{}{0em}{}
\titleformat{\subsubsection}{\normalfont}{}{0em}{}

\hypersetup{
    colorlinks=true,
    linkcolor=blue,
    citecolor=blue,
    urlcolor=blue
}

\date{}

\begin{document}

\begin{titlepage}
\centering
\vspace*{2cm}

{\LARGE\bfseries End-to-End Evaluation and Governance of an EHR-Embedded AI Agent for Clinicians \par}

\vspace{2cm}

Aaryan Shah$^{1,2,*}$, Andrew Hines$^{1}$, Alexia Downs$^{1}$, Denis Bajet$^{1}$, Paulius Mui, MD$^{3}$, Fabiano Araujo, MD, PhD$^{4}$, Laura Offutt, MD$^{3}$, Aida Rutledge, MD$^{5}$, Elizabeth Jimenez$^{3}$

\vspace{1cm}

$^{1}$Canvas Medical, San Francisco, CA, USA\\
$^{2}$Department of Biomedical Data Science, Stanford University, Stanford, CA, USA\\
$^{3}$XPC (X Primary Care)\\
$^{4}$FCA Consulting\\
$^{5}$Department of Pediatrics, University of Nevada, Reno, NV, USA\\

\vspace{1cm}

$^{*}$Corresponding author:\\
Aaryan Shah\\
Canvas Medical\\
San Francisco, CA, USA\\
aaryan.shah@canvasmedical.com

\end{titlepage}

\section*{Abstract}

Clinical AI systems require not just point-in-time evaluation but continuous governance: the ongoing practice of monitoring, evaluating, iterating, and re-evaluating performance throughout deployment. We present an end-to-end framework of governance that integrates rubric validation, live deployment feedback, technical performance monitoring, and cost tracking, with controlled experimentation gating system changes before deployment. Applied to Hyperscribe, an EHR-embedded agent that converts ambient audio into structured chart updates, twenty clinicians authored 1,646 validated rubrics across 823 cases. Seven Hyperscribe versions were evaluated through controlled experiments, with median scores improving from 84\% to 95\%. Analysis of 107 live feedback entries over three months showed feedback composition shifting from 79\% error reports and 14\% positive observations to 30\% errors and 45\% positive observations as engineering interventions resolved failures. Median processing time per audio segment was 8.1 seconds with a 99.6\% effective completion rate after retry mechanisms absorbed transient model errors. These results demonstrate that continuous, multi-channel governance of deployed clinical AI is both achievable and effective. 

\section{Introduction}

Large language models are rapidly entering clinical practice: nearly one-third of US hospitals have integrated generative AI into electronic health records (EHRs) \cite{everson2025}, and 72\% of physicians report professional use of healthcare AI \cite{ama2026}. These tools are no longer passively responding to user queries but actively making decisions, interacting with clinicians in real time, and producing structured outputs that enter the medical record. Evaluating such systems at a single point in time is necessary but insufficient. A system that performs well today may degrade after a model update, encounter a novel clinical scenario, or interact poorly with a changed workflow. What is required is governance: the continuous practice of monitoring, evaluating, iterating, and re-evaluating clinical AI systems throughout their operational lifecycle \cite{andersen2024,hassan2024}.

Governance is distinct from evaluation. Evaluation asks whether a system performs well at a point in time. Governance asks whether a system is being actively managed to perform better over time and across conditions. Governability encompasses both architectural design and evaluation infrastructure. Architecturally, it is the degree to which a system enables governance through structured output, inspectable intermediate reasoning, and constrained action spaces. Methodologically, it requires a computable performance objective that produces quantitative, repeatable scores enabling direct comparison across system versions. Governance extends beyond evaluation; it must include live user feedback, technical performance monitoring, cost tracking, and controlled experimentation that gates engineering changes before deployment. Prior work has advanced clinical AI evaluation through benchmark suites, physician-authored rubrics, and prospective deployment trials \cite{croxford2025,arora2025,tam2024,wang2025}. These contributions are valuable but address only one governance dimension. No published framework systematically governs a deployed clinical AI agent end-to-end, connecting expert judgment, live user feedback, technical monitoring, and iteration within a single continuous system \cite{andersen2024,babic2025}.

In this paper, we present three contributions. First, we describe and operationalize a governance framework integrating four dimensions: (1) rubric validation, (2) live clinician feedback, (3) technical performance monitoring, and (4) cost tracking. These dimensions inform changes in Hyperscribe's design that must pass controlled experimentation against benchmarks before deployment. We also present the full architecture of Hyperscribe, an EHR-embedded clinical AI agent designed for governability through structured outputs, explicit intermediate reasoning, a bounded action space for structured clinical actions (commands), and a computable performance objective in the form of case-specific rubrics \cite{hyperscribe_product,hyperscribe_github, canvas_sdk}. Second, we present evidence from live deployment showing that feedback-driven iteration improves quality over time, documented through thematic analysis of clinician feedback and the subsequent engineering interventions. Third, we report on the operational infrastructure supporting continuous governance, including logging and auditability for failure attribution, model inference cost tracking across pipeline stages, and technical performance monitoring for real-time reliability \cite{hyperscribe_product,hyperscribe_github}.

\section{Results}

\subsection{Feedback-driven Iteration}

Thematic analysis of 107 live deployment feedback entries identified five major themes (Figure 1A). Some entries contained multiple themes; percentages sum to greater than 100\%.

\begin{figure}[H]
\centering
\includegraphics[width=\textwidth]{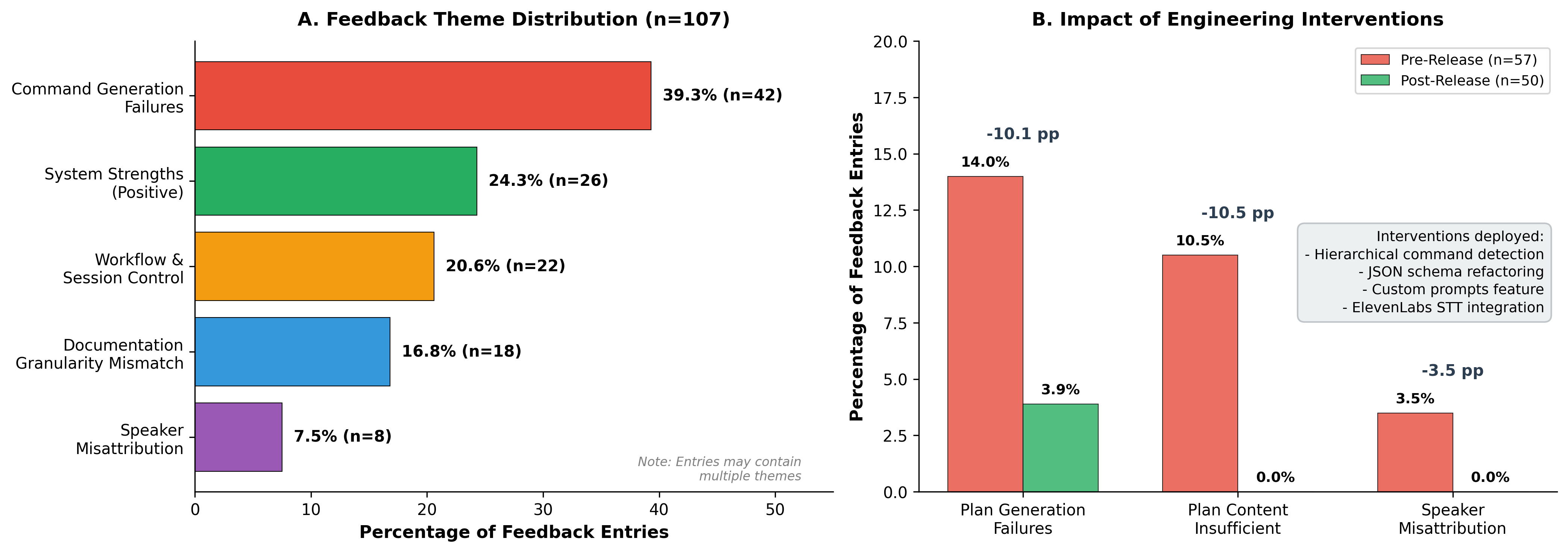}
\caption{(A) Feedback theme distribution across 107 entries. Command generation failures were the most common theme (39.3\%). (B) Impact of engineering interventions on three key failure categories, showing percentage-point reductions pre- vs.\ post-release.}
\end{figure} 

Command generation failures (failures in producing structured clinical actions for the patient's chart; 42 entries, 39.3\%) were the most common theme. Two primary concerns emerged: insufficient Assessment and Plan documentation and data completeness gaps within questionnaires. One clinician reported: ``HPI was great, the mental status exam was also quite good. Plan was not--was just one sentence. I usually do my assessment and plan together... But even without an assessment, the plans have been too brief.'' The project team employed four intervention strategies: infrastructure changes including a decision tree methodology for instruction detection and JSON schema refactoring, custom prompting for user-specific command customization, prompt engineering with improved guardrails, and customer education addressing SDK configuration issues.

Speaker misattribution (8 entries, 7.5\%) represented a smaller but clinically significant failure class. A clinician noted: ``HPI was near perfect--the only issue was it misattributed her discussion of her dad's psychiatric issues as her issues.'' Investigation traced the failure to vague speaker recognition instructions in the transcript-to-instructions prompt. The modified prompt explicitly distinguished patient, clinician, and family speakers. The team also switched speech-to-text providers to a model with stronger speaker recognition. Feedback pertaining to misattribution reduced significantly after these changes.

Documentation granularity mismatch (18 entries, 16.8\%) captured cases where clinicians found outputs either near-verbatim or overly compressed. One clinician observed: ``In both the plan and mental status, it seemed to be just almost transcribing the conversation instead of summarizing.'' The appropriate granularity varies by note type, specialty, billing requirements, and individual style. The engineering team enabled custom prompt versions allowing users to specify preferred documentation structure through guidance examples.

Workflow and session control (22 entries, 20.6\%) concerned state management and interface design. One clinician reported: ``If paused, cannot stop it from that state. I need to resume it, and then stop when it becomes available.'' The team redesigned the control interface: replacing text-based buttons with three persistent icon-based controls (Play, Pause, Stop), adding a real-time status indicator, and reorganizing the tabbed interface to surface the Transcript tab for real-time verification (Figure 2A, 2B).

\begin{figure}[H]
\centering
\includegraphics[width=\textwidth]{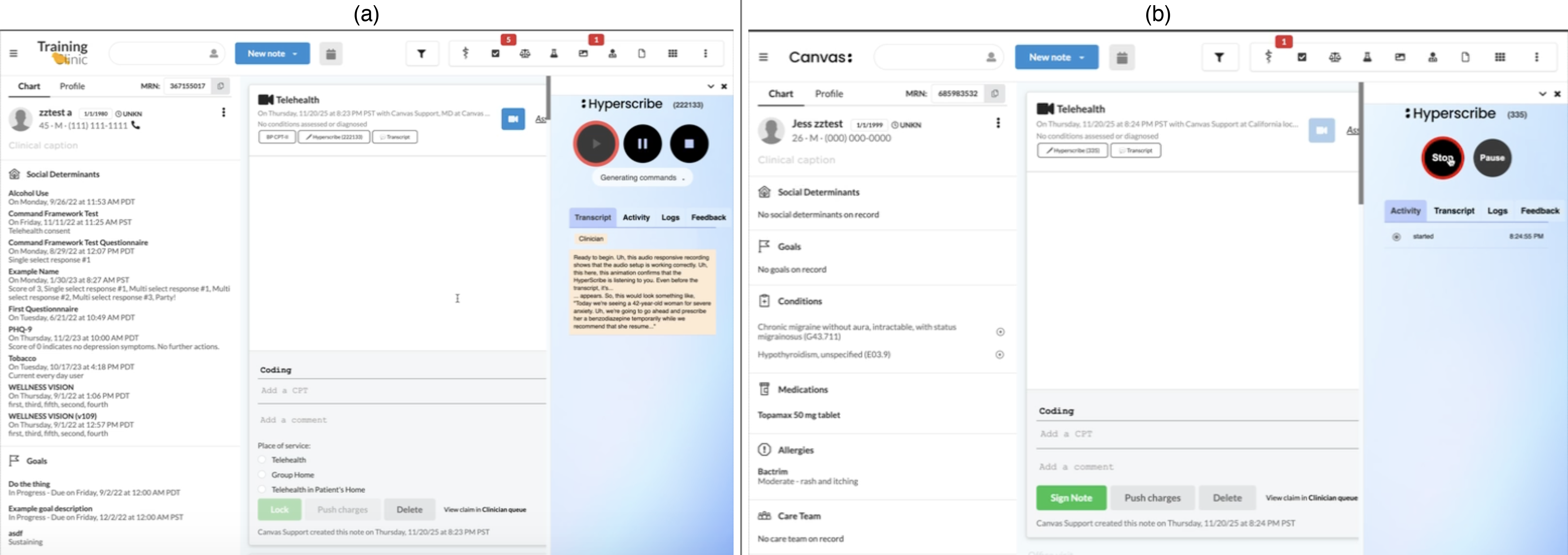}
\caption{UI redesign for session controls. (A) Original interface with text-based controls. (B) Redesigned interface with persistent icon-based controls (Play, Pause, Stop), real-time status indicator, and reorganized tabbed interface surfacing the Transcript tab.}
\end{figure}

System strengths (26 entries, 24.3\%) captured positive observations, which clustered around narrative note sections, especially History of Present Illness and Mental Status Examination. One clinician noted: ``Much better job capturing details for medication statement commands, allergies, and past medical history!'' Longitudinal improvement was explicitly recognized: after the November 2024 release, comments regarding insufficient plan details or speaker misattribution were effectively eliminated.

The temporal pattern across the three-month period is notable (Figure 1B).

Feedback composition shifted from error-dominant (78.6\% error reports in September) to balanced (30.0\% errors, 45.0\% positive observations in December). This shift is consistent with the governance loop operating as designed: identified failures are addressed through engineering interventions, and subsequent feedback reflects the improvements (Table 1).

\begin{table}[H]
\centering
\caption{Feedback Theme Analysis ($n = 107$)}
\small
\begin{tabular}{lccp{5cm}}
\toprule
\textbf{Theme} & \textbf{Count} & \textbf{\% of Total} & \textbf{Investigation Outcomes} \\
\midrule
Command Generation Failures & 42 & 39.3\% & Improvement made: 2 \newline Customization request: 2 \newline Customer Education: 2 \\
Speaker Misattribution & 8 & 7.5\% & Improvement made: 1 \\
Documentation Granularity Mismatch & 18 & 16.8\% & Customer Education: 5 \newline Customization request: 2 \\
Workflow \& Session Control & 22 & 20.6\% & Feature request: 2 \newline Improvement made: 3 \\
System Strengths (Positive) & 26 & 24.3\% & No investigation tags provisioned, affirmative responses only. \\
\bottomrule
\end{tabular}

\vspace{0.5em}
\small\textit{Note:} Some entries contained multiple themes; percentages sum to greater than 100\%.
\end{table}

\subsection{Experimental Validation}

Seven experimental versions were evaluated in controlled experiments, each producing 10 experimental notes per case (5 OpenAI, 5 Anthropic) across 823 cases (Table 2). Experiments 1--4 (control-baseline through hierarchical-detection) showed similar distributions with medians around 83--84\% and wide interquartile ranges (Q1: 50--58\%, Q3: 89--90\%). Experiments 5--7 (model-updates, model-switching, prompt-minimization) showed marked improvement: median scores increased to approximately 95\%, and the lower quartile shifted from 50--58\% to 80--81\%.

The transition between experiments 4 and 5 represents the most significant discontinuity, coinciding with foundation model updates. This suggests that model capability improvements drove quality gains more than prompt engineering or schema changes alone.

Vendor preference analysis showed Anthropic models achieving a net preference rate of +44.2 (72.1\% selected as best by clinicians) and OpenAI models achieving a net preference rate of +13.6 (56.8\%). Full results, including inter-rater reliability analysis and the convergence finding between LLM-generated and clinician-authored rubrics, are described in the companion paper \cite{paper_a}.

\begin{table}[H]
\centering
\caption{Experiment Performance Across Seven Versions}
\begin{tabular}{llcccc}
\toprule
\textbf{Experiment} & \textbf{Purpose} & \textbf{Median} & \textbf{Q1} & \textbf{Q3} & \textbf{Std Dev} \\
\midrule
Control baseline & Reference point & 84.0\% & 54.2\% & 90.0\% & 33.1 \\
Next branch & Prompts, guardrails & 83.9\% & 50.7\% & 90.0\% & 33.9 \\
JSON schema & Structured parameters & 83.6\% & 50.0\% & 89.9\% & 33.8 \\
Hierarchical detection & Section-based detection & 83.3\% & 58.5\% & 89.0\% & 30.9 \\
Model updates & Latest model versions & 94.7\% & 81.0\% & 100.0\% & 27.9 \\
Model switching & Cost reduction strategy & 94.8\% & 81.0\% & 100.0\% & 28.3 \\
Prompt minimization & Token optimization & 94.3\% & 80.0\% & 100.0\% & 28.8 \\
\bottomrule
\end{tabular}
\end{table}

\subsection{Clinician Experience with Rubric Validation}

Twenty clinicians participated in the rubric validation effort. For each case, clinicians reviewed the full transcript and medical history of the patient at the time of the visit. Then, clinicians examined 3-5 Hyperscribe outputs generated under different system versions, labeled the single best and single worst output, and authored rubric criteria encoding what Hyperscribe should produce. A second clinician independently repeated this process. This workflow is described further in the Methods below. 

Clinicians reported a median rubric validation time of 18 minutes per case (range: 5--30 minutes), with high confidence that their rubrics captured what ``correct'' should mean (median 4 out of 5) and reflected meaningful differences in AI performance. When asked how much they would trust such a system for evaluating a real clinical AI tool prior to deployment, clinicians reported a median of 8 out of 10. Half of respondents reported that the time required per case remained stable throughout the validation period, while the other half reported decreases as they gained familiarity with the validation process.

Clinicians identified several categories of clinical reasoning that proved difficult to operationalize into rubric criteria. Treatment planning decisions--including medication changes, follow-up scheduling, and imaging orders--were consistently cited as challenging to encode. One reviewer observed that ``rubrics tended to be very generic and therefore not nuanced to some situations. I found this came up mostly with treatment plans.'' The attribution problem, distinguishing clinician self-disclosure from patient history, emerged as a recurring challenge: ``There were many instances where it was unable to differentiate when a clinician shared a personal anecdote or experience, and it was captured as patient history or experience.'' Additionally, patient emotional content and sentiment proved difficult to capture in structured criteria.

The difficulty of rubric writing varied substantially by case type. Behavioral health cases with lengthy, conversational transcripts and oncology cases involving medication details, treatment roadmaps, and symptom tracking demanded more evaluation time. The central methodological challenge was achieving score separation between best and worst notes when differences were subtle. As one reviewer described: ``There were some cases where the notes I scored felt more about finding `a distinction without a difference' in order to satisfy the scoring paradigm.''

When describing what constituted a high-quality note, clinicians converged on consistent criteria: accuracy, conciseness, appropriate summarization rather than verbatim transcription, and correct placement of information in appropriate sections. Best notes ``captured new information from the transcript,'' corrected pre-existing errors, and demonstrated clinical synthesis. Worst notes exhibited omissions, errors, unnecessary repetition, or transcribed conversation verbatim rather than synthesizing it. As one reviewer summarized: ``A good note concisely and professionally captures medical information discussed during the transcript.''

\subsection{Case Provenance}

In the blinded provenance assessment, provenance judgments were collected starting in mid-November, yielding 707 classifications across 404 distinct cases (339 real-world and 70 synthetic, 49.1\% of all cases). The two clinicians who authored the accepted rubrics for each case also assessed its provenance (Table 3).

\begin{table}[H]
\centering
\caption{Case Provenance Confusion Matrix}
\begin{tabular}{lccc}
\toprule
 & \textbf{Predicted Real} & \textbf{Predicted Synthetic} & \textbf{Total} \\
\midrule
Actual Real & 379 (TP) & 218 (FN) & 597 \\
Actual Synthetic & 15 (FP) & 95 (TN) & 110 \\
\bottomrule
\end{tabular}

\vspace{0.5em}
\small Overall accuracy: 67.0\%. Balanced accuracy: 74.9\%. Clinicians correctly identified 63.5\% of real-world cases and 86.4\% of synthetic cases.
\end{table}

Across all classifications, clinicians achieved an overall accuracy of 67.0\% and a balanced accuracy of 74.9\%. Clinicians correctly identified 63.5\% of real-world cases (379 of 597 assessments) but correctly identified 86.4\% of synthetic cases (95 of 110 assessments). This asymmetry suggests that synthetic cases contained more detectable artifacts than real cases.

Clinicians developed heuristics for distinguishing case origin. Real case indicators centered on conversational authenticity: interruptions, colloquialisms, cognitive shifting between topics, breaks in speech patterns, and clinician self-disclosure. As one reviewer noted, real transcripts featured ``breaks in sentence structure, cognitive shifting, `normal' flows of speech and cognition.'' Synthetic case indicators included unnaturally polished language, perfect conversational turn-taking, and structural anomalies. Reviewers described synthetic transcripts as having ``serve and volley conversations--no talking over each other, directed to the point responses'' and ``unnaturally polished language, perfectly ordered turn-taking, unrealistic patient phrasing.'' Technical artifacts such as fictitious medications or mismatched demographics between transcript and chart also signaled synthetic provenance. The distinction between real-world cases and synthetic cases is informative for governance: it confirms that while synthetic cases are useful for coverage of underrepresented clinical scenarios, they cannot yet fully subtitute for real clinical encounters. 
\subsection{Technical Performance}

Median end-to-end processing time, from instruction detection through command generation). Per-stage analysis identified instruction detection as the sequential bottleneck at median 5.9 seconds (P95: 16.3s), reflecting the complex nature of identifying clinical instructions and intent from transcript. Audio transcription and speaker diarization tasks were comparatively fast (median 1.1s and 1.6s, respectively).

Failure rates during experiments were also of interest. Across all system versions, 6.6\% of note generation attempts encountered at least one error. Error categorization revealed two dominant failure modes: unexpected LLM output format and parameter type mismatches. The retry mechanisms effectively absorbed these transient failures (JSON retries for schema-nonconforming outputs), recovering 94.2\% of errors and yielding an effective completion rate of 99.6\% across all experiments.

\subsection{Cost}

\begin{table}[H]              
  \centering                                               
  \caption{Cost Summary of Governance Infrastructure}
  \small                                                
  \begin{tabular}{llcc}
  \toprule                  
  \textbf{Category} & \textbf{Component} & \textbf{Compute Cost} & \textbf{Clinician Cost} \\             
  \midrule                   
  Case construction (static) & Real-world (736 cases) & \$317 & -- \\                                     
   & Synthetic (87 cases) & \$3 & -- \\                                            
  \midrule                                                   
  Rubric generation (static) & Clinician-authored (1,646) & -- & 919 hours \\
   & LLM-generated (823) & \$14 & -- \\                 
  \midrule                                                                                                
  Experimentation (across 7 versions) & Note generation (57k notes) & \$15,000 & -- \\
   & Rubric scoring & \$10,000 & -- \\                                                                    
  \midrule                                                                         
  \textbf{Total} & & \textbf{\$25,334} & \textbf{919 hours} \\                                            
  \bottomrule                
  \end{tabular}                                                                                           
                                                                                                          
  \vspace{0.5em}                                     
  \small\textit{Note:} Static costs are incurred once. Experimentation costs are approximately \$3,600 per
   version.                                                                        
  \end{table}     

Case construction cost approximately \$320 for the 823-case benchmark: \$317 for 736 real-world cases extracted from live clinical audio via a multi-step pipeline (approximately \$0.43 per case) and \$3 for 87 synthetic cases generated using OpenAI's o3 reasoning model.

Clinician validation for the 823 cases involved 919 total hours across 20 evaluators across a five-month span, including onboarding and training hours. This yielded a total of 3,108 validated rubrics (of which 2,469 were used in experimentation) at a mean effort of 17.7 minutes per accepted rubric. This effort, however, encompasses the full validation workflow of reviewing case materials, identifying best and worst notes, authoring rubric criteria, and verifying rubrics achieved score separation. Clinicians self-reported a median of 18 minutes per case (range 5--30 minutes). Half of these clinician annotators reported stable time requirements for this process per case throughout the validation period, but the other half reported that this decreases with more experience crafting rubrics and understanding the workflow.

823 LLM-generated rubrics (one per case) were produced by o3, producing around 2.05 million total input tokens and 1.23 million output tokens across these rubrics. Assuming a 5x reasoning overhead for reasoning tokens that are billed at the output rate (o3 pricing: \$2/M input, \$8/M output), the total cost was approximately \$14 for all LLM rubric generation for the dataset. This represents approximately \$0.02 per rubric, compared with approximately \$29.50 per clinician-authored rubric, a 1000x difference. The quality trade-off between these paths is characterized in the convergence analysis previously and a detailed cost analysis of the full evaluation infrastructure is reported in \cite{paper_a}.

For experimental inference, the per-note consumption ranged between 37,000 and 44,000 prompt tokens and 10,000 and 13,000 output tokens across Anthropic and OpenAI implementations, respectively. Total experimental inference cost across seven versions was approximately \$25,000\$10,000 OpenAI), encompassing both note generation and rubric scoring across approximately 57,000 notes (823 cases, 10 notes each case, seven Hyperscribe versions). 

In later experiment versions, where cost reduction was of primary interest, two versions reduced cost burden. The model-switching experiment (experiment 6) routed simpler pipeline stages to smaller models, which kept token counts comparable, but the price differential yielded an estimated 20--30\% cost reduction depending on vendor without measurable quality reduction (median rubric scores: 94.8\% vs.\ 94.7\%). Separately, prompt optimization efforts in experiment 7 reduced per-note prompt tokens by 4--6\% relative to baseline for both vendors.

These costs divide into static and growing components. Static costs, incurred once to build the         
evaluation infrastructure, include case construction (\$320) and LLM rubric generation (\$14). Growing
costs, incurred per system version, include experimental inference at approximately \$3,600 per version.
The dominant investment was clinician time: 919 hours across 20 evaluators for rubric authorship, representing the time spent by experts encoding their knowledge into rubrics to evaluate Hyperscribe. 

\section{Discussion}

\subsection{Prior Work}

The governance of deployed clinical AI systems has received substantial theoretical attention, yet empirical implementations remain rare. Hussein et al. systematically reviewed 35 governance frameworks and proposed a maturity model for assessing organizational AI readiness \cite{hussein2026governance}. Wells et al. introduced the FAIR-AI framework to guide health systems through pre- and post-implementation evaluation \cite{wells2025fairai}. These contributions establish important conceptual foundations, but lack operational data from deployed governance efforts. Our framework not only provides empirical evidence that continuous, multi-channel governance is feasible, but also introduces design decisions in clinical AI to ensure that tools are governable. The integration of controlled experimentation, rubric validation, live clinician feedback, technical monitoring, and cost tracking within a single, continuous system, has not, to our knowledge, been demonstrated in the literature. 

The literature on ambient AI scribe deployment has focused on efficiency and clinician experience rather than clinical output quality. Afshar et al. described a pragmatic trial
playbook for monitoring ambient AI across 20 providers and 8,527 encounters, measuring documentation time and clinician burnout but not the accuracy or completeness of generated content \cite{afshar2025pragmatic}. The NEJM AI randomized trial comparing two ambient scribe platforms across 238 physicians, similarly prioritized efficiency outcomes and noted occasional inaccuracies without a systematic method to characterize or resolve them \cite{nejmrct2025ambient}. However, this measurement gap is consequential: Asgari et al. demonstrated that even at a 1.47\% hallucination rate, nearly half of hallucinated content was clinically major \cite{asgari2025hallucination}. Quality monitoring cannot be deferred to post-hoc audits or user complaints alone. Our rubric-based validation provides the missing quality dimension, and our governance architecture ensures that detected issues trigger engineering interventions whose effects are subsequently re-measured in controlled experiments.

\subsection{Evidence for a Functioning, Cost-Effective Governance System}

The temporal shift in feedback composition, from 78.6\% errors and 14.3\% positive observations in September to 30.0\% errors and 45.0\% positive observations in December, is evidence that the multi-channel governance loop functions in a scalable way. The system connects results to identified failures, such as feedback themes leading to specific engineering interventions and new feedback to reflect the resolution of previous failures or regressions.

This connection is traceable across multiple themes. Speaker misattribution feedback led to a prompt rewrite explicitly distinguishing patient, clinician, and family speakers, combined with a switch to a speech-to-text provider with stronger speaker recognition. Following these changes, misattribution feedback dropped to two entries. Command generation complaints about plan brevity led to infrastructure changes (decision tree methodology for instruction detection, JSON schema refactoring) and custom prompting capabilities. After the November 2025 release incorporating these changes, feedback regarding insufficient plan details was effectively eliminated.

The shift from error reports to feature requests is itself a qualitative governance signal. When users move from reporting failures to requesting new capabilities, it suggests that the baseline functionality has become reliable enough that attention shifts to expansion. This is distinct from the absence of feedback, which could indicate disengagement. The persistence of feedback volume alongside the compositional shift suggests continued user engagement with an improving system.

\subsection{Governability as a Requirement}

Hyperscribe's architecture was designed for governability. Structured outputs make failures attributable to specific pipeline stages. When a command generation failure occurs, audit logs trace whether it originated in instruction detection (wrong intent), parameter filling (wrong field values), or command construction (wrong chart interaction). This ensures that non-deterministic models with black-box reasoning can be investigated in a targeted manner. Explicit intermediate reasoning enables targeted fixes with bounded regression risk. When speaker misattribution was identified, the team traced the failure to the transcript-to-instructions stage and modified only that stage's prompt, without needing to re-validate or re-engineer other pipeline stages.

The EHR-bounded action space further constrains risk because the system can only produce predefined command types validated against the patient's chart, avoiding the introduction of entirely novel error categories. Lastly, the computable performance objective, in the form of case-specific rubrics, provides the measurement basis for controlled experiment, enabling direct comparison.

The graduated context injection, from raw audio at stage 1 to full patient chart at stage 4, serves both clinical and governance purposes. Clinically, it prevents early-stage overweighting of chart data. For governability, it creates clean boundaries for failure attribution: if a Diagnosis command contains an error, logs reveal whether the error entered at instruction detection, parameter filling, or command construction. These are decisions made during system design, not retrofitted after deployment, that determine whether failures can be attributed, investigated, and resolved efficiently. Clinical AI systems that produce free-form text from opaque reasoning struggle to offer these attribution and isolation properties.

\subsection{Operational Lessons from Feedback-driven Iteration}

Three operational patterns emerged from the feedback-experimentation-deployment pipeline that may generalize beyond this specific system.

First, prompt-level fixes were fast and effective for well-scoped failures. Speaker misattribution was resolved within one release cycle through a targeted prompt rewrite. This speed of iteration is a direct consequence of intermediate reasoning because when it is feasible to localize a failure to a specific stage (and therefore, a specific set of functions and prompts), the fix is also localized.

Second, infrastructure changes required more effort but had broader impact. The decision tree methodology for instruction detection and the JSON schema refactoring addressed multiple failure modes simultaneously. The UI redesign for session controls addressed an entire class of workflow problems that were invisible to the rubric-based evaluation.

Third, given the noted diversity of physician preference in documentation, there is no universally correct level of detail, only clinician-specific and organization-specific expectations. Here, custom prompting was the resolution, enabling users to specify details about their practice setting, documentation philosophy, and exact syntactic preferences. Similarly, some feedback traced to customer education rather than code changes to inform users for how the system worked and why, for example, Hyperscribe reorganizes information rather than producing free-text across the note. The taxonomy of intervention types provided in this paper may apply to other clinical AI systems.

Beyond live feedback, the governance framework's evaluation coverage depends on the diversity of cases when testing the system. From a governance perspective, the relevant question is not whether synthetic cases are indistinguishable from real cases, but whether they produce evaluation signals informative for system improvement. The ability to generate synthetic cases across clinical domains and edge cases rare in the real-world corpus is a governance capability that enables cost-effective, continuous evaluation.

\subsection{Cost and Technical Performance in Governance}

Cost tracking and technical performance monitoring support governance efforts by ensuring that quality improvements are sustainable and viable operationally.

The cost structure reveals a 1000x difference between clinician-authored and LLM-generated rubrics, confirming that expert availability is the primary constraint on evaluation coverage. Furthermore, cost optimization is supported by the infrastructure as seen in the model switching and prompt minimization strategies that reduced inference cost burden--all part of the same experimental framework used for clinical quality assessment.

The technical monitoring infrastructure, in our opinion, is a significant governance contribution. The logging architecture that supports consolidated summaries, all the way down to the granular prompt-response histories, enables failure attribution at scale. Alongside the retry mechanism and structured error records, this infrastructure makes it easy to distinguish between model failures, data configuration errors, or transient API call issues.

Lastly, we demonstrated that the real-time latency offered by Hyperscribe is acceptable clinically, while instruction detection and questionnaire processing represent optimization targets. These targets will become increasingly important as reasoning models, which are currently excluded from the real-time pipeline due to prohibitive and unpredictable processing times, continue to improve in speed. The governance infrastructure described here provides the measurement framework to evaluate that tradeoff: when reasoning models become fast enough to consider, the same per-stage latency monitoring and experimental comparison pipeline can quantify whether their quality advantages justify their latency cost in a clinical setting.

\subsection{Limitations}

We acknowledge limitations to methods in this paper.

First, the governance framework is validated on a single system (Hyperscribe) within a single EHR (Canvas Medical). Hyperscribe's architecture is different from other scribes and other clinical AI systems operating with free-form text or outside an EHR. However, the governance framework is not architecture-dependent and this loop can be implemented with different mechanisms for failure attribution and targeted intervention. Furthermore, we believe that a governance-forward architecture is vital in clinical settings for safety.

Second, the live feedback sample of 107 entries over three months from early adopter practices is limited in scale and may overrepresent engaged power users who are most invested in providing detailed feedback. The thematic analysis identifies five failure categories that are not exhaustive. However, the validity of the feedback-to-intervention pipeline does not depend on the exhaustiveness of the feedback corpus. The question addressed by this paper is whether the governance loop functions, not just whether the feedback themes represent the complete distribution of possible failures. As deployment scales, the feedback corpus will grow and shift as engineering interventions accrue.

\section{Conclusion}

This paper presents the first end-to-end governance framework for a deployed EHR-embedded clinical AI agent, integrating rubric validation, live deployment feedback, technical performance monitoring, and cost tracking within a single continuous system, where controlled experimentation gates iterative changes before deployment. Prior work has addressed individual governance dimensions in isolation but none operationalize these channels together for a production-grade, clinical AI system. The framework's four dimensions proved complementary in practice. Rubric-based evaluation detected an approximately 11-percentage-point improvement in median clinical quality across seven system versions. Live clinician feedback identified five distinct failure themes, each leading to targeted engineering interventions validated through controlled experimentation before release. The temporal shift from error-dominant feedback (78.6\% errors, 14\% positive) to balanced feedback (30.0\% errors, 45.0\% positive) over three months provides evidence that the governance loop drives measurable, user-visible improvement. Operationally, the system maintained a median end-to-end cycle latency of 8.1 seconds with retry mechanisms absorbing transient model failures to maintain reliable outputs. Cost optimization strategies reduced inference costs by 20--30\% without measurable quality degradation, as measured by the rubric validation effort. 

Hyperscribe's structured outputs, intermediate reasoning, constrained action space, and computable performance objective make governance tractable: failures are attributable to specific pipeline stages, fixes can be targeted without system-wide re-validation, and rubric-based scoring enables direct version comparison through controlled experimentation.These properties suggest that clinical AI systems should be designed for governability from conception. As clinical AI systems become more deeply embedded in patient care, the question shifts from whether systems can perform well to whether they can be governed to perform well. This paper demonstrates that continuous governance is achievable when multiple channels operate in concert to improve usability and performance of clinical AI tools and systems.

\section{Methods}

\subsection{Governance Framework Overview}

Our governance framework integrates four interconnected activities: (1) rubric validation, (2) live clinician feedback integration, (3) technical performance monitoring, and (4) cost tracking (Figure 3). Controlled experimentation using the rubric-based benchmark connects these dimensions such that changes to Hyperscribe's architecture informed by any dimension must be tested against the full benchmark before deployment to confirm expected performance in real-world settings.

\begin{figure}[H]
\centering
\includegraphics[width=\textwidth]{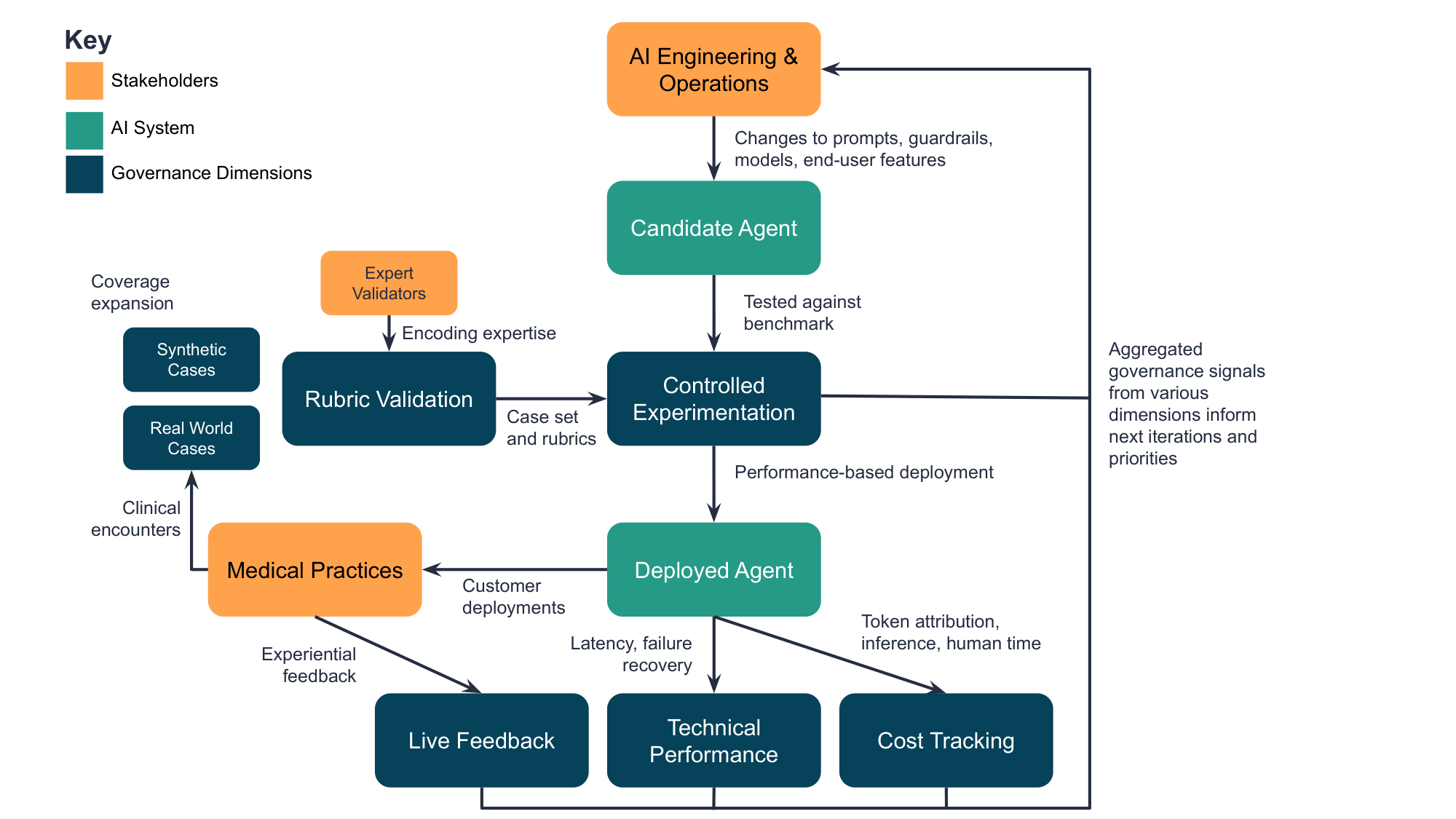}
\caption{Governance framework integrating rubric validation, live clinician feedback, technical performance monitoring, and cost tracking, with controlled experimentation gating iterative changes before deployment. Signals from the four dimensions inform priorities. Candidate system versions are tested against the full rubric benchmark before deployment. The deployed agent generates new signals, continuing the cycle.}
\end{figure}

These activities operate as a continuous cycle rather than as independent evaluation efforts \cite{andersen2024,babic2025}. Live clinician feedback surfaces failure modes encountered during deployed use. The engineering team investigates and implements targeted changes. The modified system is re-evaluated by generating experimental notes across evaluation cases and scoring them against the full rubric set, comparing resultant scores against previous system versions to verify improvements and detect regressions. Technical monitoring confirms that changes do not degrade operational performance, and cost tracking connects the evaluation effort to economic sustainability. The updated version is deployed, generating new feedback and restarting the cycle.

The rationale for integrating multiple channels is that each operates on different data, timescales, and sensitivities. Rubric-based evaluation is systematic and repeatable but operates on static case representations and cannot capture live workflow use. Live feedback captures these real-use phenomena but is episodic and cannot uniformly compare system versions. Cost tracking provides the economic context for sustained governance, and technical monitoring addresses real-time operational reliability. The architectural properties described in the System Architecture section \textemdash \space structured outputs, explicit intermediate reasoning, a bounded action space, and a computable performance objective \textemdash \space are what make this multi-channel governance tractable.

\subsection{System Architecture}

Hyperscribe is an EHR-embedded clinical agent developed by Canvas Medical that converts ambient clinical audio and patient context into structured chart updates \cite{hyperscribe_product}. Rather than producing a single free-form note from a complete transcript, Hyperscribe transforms audio into increasingly structured representations through chunkwise processing (Figure 4).

\begin{figure}[H]
\centering
\includegraphics[width=\textwidth]{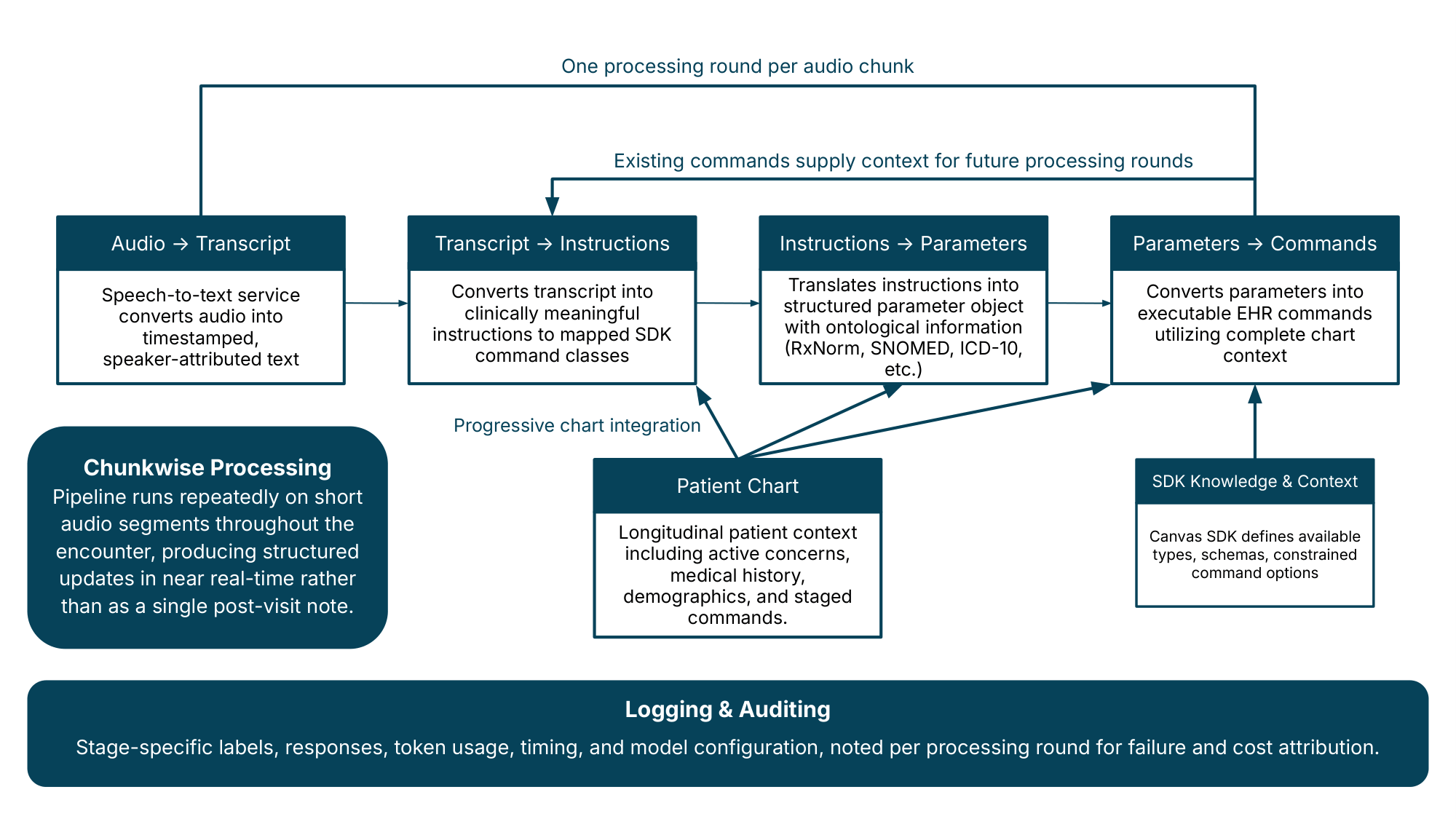}
\caption{Hyperscribe processing pipeline. Four stages form a cyclical process: audio to transcript, transcript to instructions, instructions to parameters, and parameters to commands (structured clinical actions). Patient chart context is progressively integrated across stages and commands from previous processing rounds supply context for future processing rounds. All stages are logged for failure and cost attribution.}
\end{figure} 

Each processing round can modify outputs from any preceding round through ``quadratic revision'': as the number of processing rounds grows, the number of potential revisions increases quadratically, because each new round can update documentation produced by any prior round. In practice, information introduced late in an encounter--such as a diagnosis that recontextualizes earlier symptoms--propagates backward through the documentation without requiring a complete regeneration.

The pipeline comprises four stages forming a cyclical process. The length of the processing round (``chunk'') is a parameter that can be adjusted in governance and experimentation efforts with Hyperscribe.

Stage 1: Audio to transcript. Hyperscribe records clinical audio and transcribes it into timestamped utterances with speaker labels. As a model-agnostic tool, it supports multiple speech-to-text services. A transcript tail from the previous audio chunk maintains continuity across segments.

Stage 2: Transcript to instructions. The speaker-attributed transcript is converted into instructions, which are clinically meaningful intents mapped to discrete EHR actions. Each instruction carries a unique identifier, detection order, instruction type mapped to a supported command class, and a free-text information field. The action space is dynamically constrained by patient context. Hyperscribe uses a complexity-dependent detection strategy: simple encounters use a single LLM call, while complex encounters use hierarchical detection by note section to reduce context size.

Stage 3: Instructions to parameters. Each instruction is translated into a structured parameter object defined by the target command type. Clinical ontology services map free-text language to standardized codes: ICD-10 for diagnoses, RxNorm for medications, and SNOMED CT for clinical findings. Minimal patient demographics are injected to ground interpretation, but the broader chart history is deliberately excluded at this stage to prevent hallucination from chart data rather than the current conversation.

Stage 4: Parameters to commands. Parameter objects are converted into executable commands with the richest chart context. In Canvas's architecture, a ``command'' is a structured clinical action corresponding to a discrete update to the patient's chart: for example, the Diagnose command adds a condition to the problem list, Prescribe creates a medication order, and Assess and Plan generate clinical reasoning documentation \cite{canvas_sdk}. Patient context enters the pipeline incrementally--raw audio at stage 1, speaker-attributed transcript at stage 2, minimal demographics at stage 3, and the full patient chart here. This graduated introduction prevents early-stage anchoring on chart data. The command stage also retrieves commands from other sources: intake questionnaires, vitals entered by medical assistants, and pre-populated note templates.

All LLM interactions across all four stages are logged with stage-specific labels. When audit logging is enabled, the system records prompts, responses, token usage, timing, and model configuration. These logs enable failure attribution to specific pipeline stages and support the cost analysis described in the Cost Tracking section.

Two properties distinguish Hyperscribe from conventional ambient scribes. First, it processes audio in near real-time chunks rather than producing a monolithic post-visit note. Second, it integrates EHR context into reasoning at every processing round, enabling chart-aware documentation that evolves as the encounter progresses. These structured, inspectable intermediate steps make governance tractable regardless of underlying model quality. Hyperscribe is open-source and publicly available \cite{hyperscribe_github}.

The architecture reflects four design considerations that facilitate governance (Figure 5).

\begin{figure}[t]
\centering
\includegraphics[width=0.75\textwidth]{figureB5.png}
\caption{Four design principles enabling governability: structured outputs, explicit intermediate reasoning, EHR-bounded action space, and computable performance objective.}
\end{figure} First, all pipeline stages produce structured, schema-defined objects rather than free-form text, enabling automated validation and targeted debugging when failures occur. Second, the pipeline maintains explicit intermediate reasoning layers--transcripts, instructions, parameters, and commands--each of which are independently traceable to identify where failures or regressions occurred. Third, the system operates within an EHR-bounded action space, constrained to predefined command types with chart-aware validation, which limits outputs to actions the EHR can execute. Fourth, the system should define a computable performance objective: case-specific rubrics that produce quantitative scores for any system output, enabling direct version-to-version comparison and establishing the measurement basis for the controlled experimentation gate.

\subsection{Live Qualitative Feedback}

Live deployment users of Hyperscribe at practices using the Canvas Medical EHR platform provided feedback during real patient encounters.

Feedback was collected through in-workflow mechanisms allowing free-text comments during or after Hyperscribe encounters. Over 107 entries spanning approximately three months (September to December 2025), each entry was linked to internal investigation records. The project team categorized investigation outcomes using a structured taxonomy with six categories (Table 4).

\begin{table}[H]
\centering
\caption{Investigation Outcome Taxonomy}
\small
\begin{tabular}{lll}
\toprule
\textbf{Category} & \textbf{Definition} & \textbf{Example} \\
\midrule
Improvement made & Engineering fix deployed & Modified prompt to distinguish \\
 & (prompt or code change) & past vs.\ current medication orders \\
Customer input & Issue ambiguous; requires & ``Please confirm expected \\
requested & clarification from clinician & behavior for this note type'' \\
Customer education & System working as designed; & Clinician expected automatic \\
 & expectation mismatch & population of field that requires \\
 & & manual trigger \\
Feature request & Requires new capability; & ``Ability to resume recording \\
created & tracked in product backlog & after Stop button'' \\
Investigation issue & Complex technical issue & Cross-patient data contamination \\
created & requiring extended research & (root cause unclear) \\
Customization & Organization-specific & Custom template for \\
request & configuration needed & specialty-specific note structure \\
\bottomrule
\end{tabular}
\end{table}

We performed thematic analysis on the 107 entries to identify recurring failure patterns and their relationship to engineering interventions. Temporal analysis tracked the shift in feedback composition across the three-month period.

\subsection{Rubric Validation Methodology}

Our rubric validation effort combines real (89.4\%) and synthetic (10.6\%)  clinical cases with LLM-generated and clinician-authored, case-specific rubrics (Figure 6).

\begin{figure}[H]
\centering
\includegraphics[width=\textwidth]{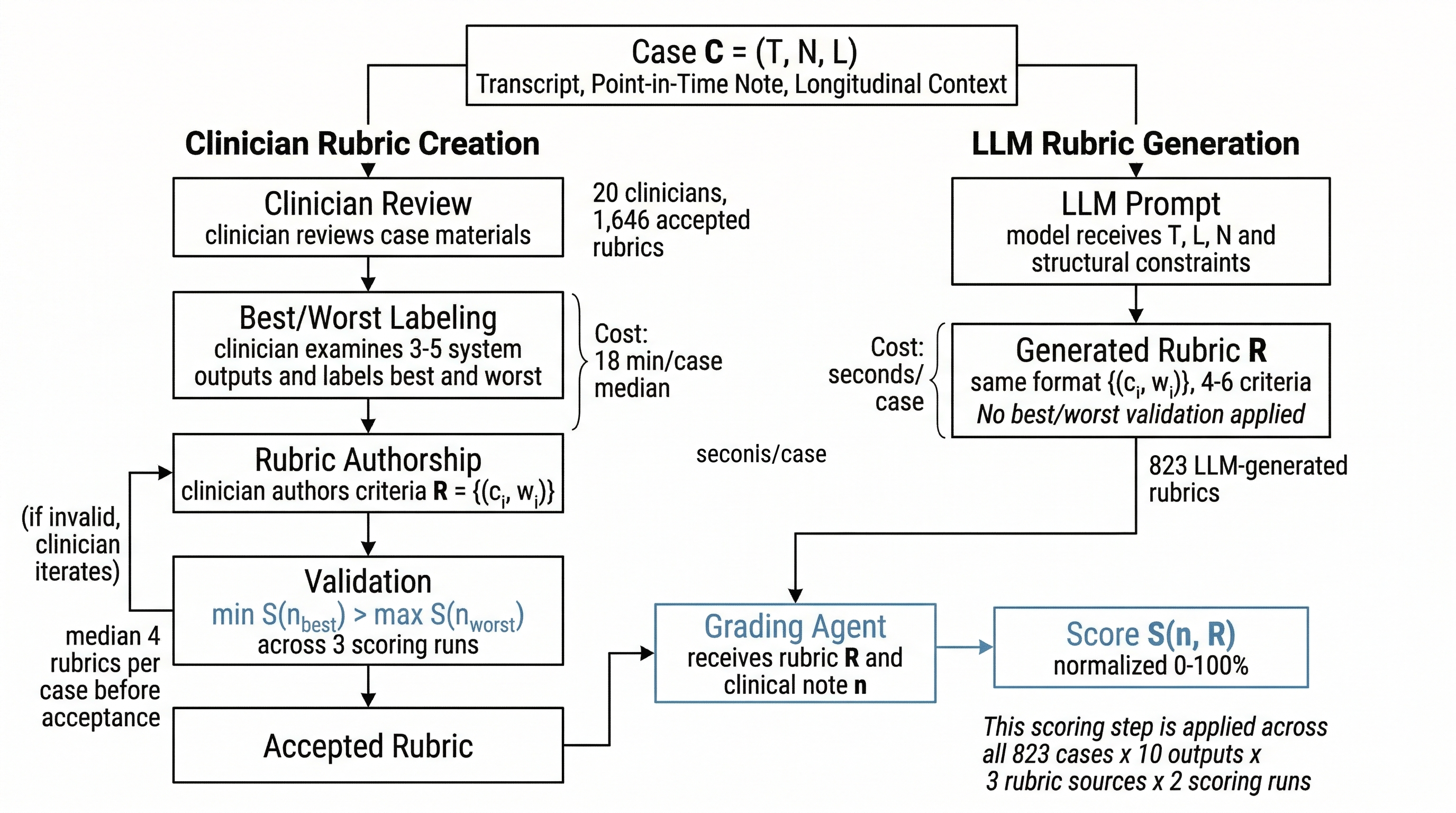}
\caption{Rubric methodology workflow (reproduced from companion paper). Two parallel paths for rubric creation (clinician-authored and LLM-generated) converge at a shared scoring agent. Clinician path includes case review, best/worst labeling, rubric authorship, and validation. LLM path generates rubrics from the same case inputs without best/worst validation.}
\end{figure} What constitutes a correct clinical note varies with the patient's history, the clinical domain, the encounter type, and the documentation conventions of the practice setting--a single universal rubric cannot accommodate this variability. Instead, through this validation effort, each case is linked to its own set of rubrics that encode what a clinician or LLM with full context considers as appropriate documentation. The methodology is described in full in the companion paper, a released dataset through PhysioNet, and GitHub repository for dataset usage; we summarize key elements in this section \cite{paper_a,physionet,dataset_usage}.

\begin{table}[t]
\centering
\caption{Dataset Characteristics}
\begin{tabular}{ll}
\toprule
\textbf{Characteristic} & \textbf{Value} \\
\midrule
Total cases & 823 \\
Real-world / Synthetic & 736 (89.4\%) / 87 (10.6\%) \\
Unique patients (real-world) & 168 \\
Median transcript length & 419 words, 37 turns (range: 28--9,402 words) \\
Demographics available & 100\% \\
Point-in-time notes & 90.4\% \\
Patient age & Median 60, mean 55.8, range 9--91 \\
Patient sex & Female 65.4\%, Male 34.1\%, not specified 0.5\% \\
Specialties represented & 35 categories \\
Encounter length & Short 46.7\%, Medium 40.8\%, Long 12.5\% \\
Problem count & Single 56.7\%, Multi 43.3\% \\
Acuity & Low 60.6\%, Moderate 35.0\%, High 4.4\% \\
Accepted rubrics & 1,646 from 20 clinicians \\
Total rubrics created & 5,797 (3,060 achieving validation; 1,646 selected for use) \\
Median criteria per rubric & 5 \\
Median rubrics per case & 4 \\
\bottomrule
\end{tabular}
\end{table}

Dataset. The benchmark (summarized in Table 5 below) comprises 823 cases: 736 real-world cases (89.4\%) from live clinical encounters across 168 unique patients, plus 87 synthetic cases (10.6\%). Real-world cases were de-identified per HIPAA Safe Harbor requirements. A case $C = (T, N, L)$ consists of a transcript $T$, a point-in-time note $N$, and longitudinal patient context $L$.

Rubric creation. For each case, two clinicians independently constructed rubrics encoding expectations for appropriate documentation behavior. A rubric $R = \{(c_i, w_i) \mid i = 1, \ldots, k\}$ defines weighted criteria where $c_i$ is a natural-language documentation requirement and $w_i$ reflects clinical importance. In developing the 1,646 accepted rubrics, 20 clinician evaluators created 5,797 total rubrics (median 4 per case, mean 7.04). A rubric was accepted and considered valid if and only if the minimum score of the clinician-selected best note exceeded the maximum score of the worst note across three independent scoring runs for each note by an LLM scoring agent.

Scoring. For a clinical note $n$ scored against rubric $R$, the normalized score $S(n, R)$ yields a percentage on a 0--100\% scale, where criterion satisfaction is weighted by clinical importance \cite{paper_a}.

\subsection{Controlled Experimentation}

Each system change to Hyperscribe was evaluated through controlled experimentation before deployment. A candidate system version was run against the full 823-case benchmark, generating 10 notes per case (five from each foundation model provider, OpenAI and Anthropic). Each note was scored against all associated rubrics using the same LLM-based scoring agent, prodcuing comparable scores across versions.

Seven experimental versions were evaluated, each representing a distinct combination of model updates, parameter tuning, and guardrail strategies informed by the governance dimensions. Version-to-version comparison of rubric scores determined whether changes improved, maintained, or degraded clinical quality. This gating mechanism ensured that no engineering change was deployed without quantitative evidence of its projected clinical impact. 

\subsection{Clinician Survey and Case Provenance Assessment}

Following the completion of rubric validation, we administered a 45-question mixed-methods survey to the 20 clinician annotators who participated in the rubric authorship and validation process. The survey instrument spanned quantitative Likert-scale measures and open-ended qualitative prompts organized across five domains: (1) time and cognitive load, (2) rubric creation workflows and evolution, (3) clinical judgment, (4) case provenance, and (5) note quality definitions. This survey was distinct from the live deployment feedback described in the Live Qualitative Feedback section; it specifically assessed the clinician experience of constructing and validating rubrics rather than the experience of using Hyperscribe in clinical practice.

Additionally, to assess the realism of synthetic cases in the dataset, we conducted a blinded provenance perception task. During the rubric validation process, clinicians were not informed of case origin and were asked to classify each case as either real or synthetic. These judgments were recorded alongside rubric scores and used to assess whether synthetic cases produce evaluation signals comparable to real-world cases--a question relevant to the scalability of governance, since synthetic cases could enable test case generation across clinical domains, populations, and scenarios that are rare or absent in the real-world corpus.

\subsection{Technical Performance Monitoring}

Hyperscribe operates in real time during live clinical encounters, executing many LLM calls, chart queries, and command-generation steps per processing round. All pipeline stages use non-reasoning models (Sonnet 4.5, GPT-4.1) to maintain predictable latency; reasoning models were deliberately excluded from the real-time pipeline despite potential quality advantages. Each LLM call incorporates a two-layer retry mechanism: up to three HTTP-level retries for transient API failures, and up to three JSON-level retries when the model returns syntactically invalid or schema-nonconforming output.

The monitoring infrastructure implements a tiered logging architecture including per-cycle summaries of duration and token counts, per-stage request-response logs, and per-instruction conversation histories. A post-session audit capability also enables granularity for investigations of feedback received during live deployments.

We report on four performance dimensions: (1) per-stage latency from timestamped production logs, (2) end-to-end cycle times, (3) failure rates per vendor and experiment, and (4) error categorization from structured error records.

\subsection{Cost Tracking}

We track four cost components: (1) case construction costs for building the evaluation dataset; (2) clinician validation costs derived from weekly hour logs submitted by 20 evaluators across each stage of the validation effort (case review, provenance classification, best-worst ranking, rubric authorship, scoring review); (3) LLM rubric generation costs; and (4) model inference costs for note generation throughout the experiments.

For all model inference costs in (1), (3), and (4), costs were calculated based on per-call token counts and confirmed by API usage and billing records.

\section{Data and Source Code Availability}

The de-identified dataset is available through PhysioNet \cite{physionet}. Companion scripts for data loading, scoring, and LLM rubric generation are available in the dataset usage repository \cite{dataset_usage}. The Hyperscribe source code is publicly available \cite{hyperscribe_github}.

\section*{Acknowledgments}

We thank Paulius Mui and the XPC (X Primary Care) community for organizing and collaborating with the clinicians who authored and validated the evaluation rubrics for this dataset. We thank the following clinicians for their rubric contributions accompanying the clinical scenarios: Amy Close, Fabiano Araujo, Laura Offutt, Sue StPierre, Aida Rutledge, Kevin Stephens, Elizabeth Jimenez, Ganga Nadella, Denton Shanks, Ann Pongsakul, Renee Dua, Caryn Mark, Natalie Freels, Yimdriuska Magan, Timothy Tsai, Rachel Menon, Alexandra Ristow, Casey Kirkhart, and Kevin Volkema.

We also thank the clinician users at Canvas Medical customer organizations who used Hyperscribe during its experimental phase and provided feedback that informed system improvements.

\section*{Competing Interests}

AS, AH, AD, and DB are employees or contractors of Canvas Medical and developed the Hyperscribe platform from which clinical encounter data was collected and organized. PM, FA, LO, AR, and EJ served as clinician evaluators for the rubric validation effort along with contributing to the copy of this research. Canvas Medical funded the clinician annotation effort that produced the evaluation rubrics. The dataset was prepared for public release to support independent research on clinical documentation evaluation. The authors have no financial relationships with LLM model providers beyond standard API service agreements.

\section*{Author Contributions}

AS conceived and designed the study, developed the evaluation infrastructure, conducted the analysis, and drafted the manuscript. AH provided strategic direction, reviewed methodology and results, and contributed to manuscript revision. AD contributed to system development, data pipeline engineering, and manuscript review. DB contributed to system development and experimental infrastructure. PM organized and coordinated the clinician evaluation panel. FA, LO, AR, and EJ authored and validated evaluation rubrics, participated in the clinician survey, and reviewed the manuscript. All authors approved the final version.

\section*{Funding}

This work was funded by Canvas Medical. No external funding was received.

\section*{Ethics Statement}

All clinical data was de-identified in accordance with HIPAA Safe Harbor standards. Institutional review board (IRB) approval was not required.



\begin{thebibliography}{18}

\bibitem{everson2025}
Everson J, Nong P, Richwine C.
Uptake of generative AI integrated with electronic health records in US hospitals.
\textit{JAMA Netw Open}. 2025;8;(12):e2549463.
doi:10.1001/jamanetworkopen.2025.49463

\bibitem{ama2026}
American Medical Association. 
Physician survey on augmented intelligence. Mar 2026. Available from \url{www.ama-assn.org/system/files/physician-ai-sentiment-report.pdf}

\bibitem{andersen2024}
Andersen ES, Hardt J, Nielsen C, et al.
Monitoring performance of clinical artificial intelligence in health care: a scoping review.
\textit{JBI Evidence Synthesis}. 2024;22(12):2362--2419.
doi:10.11124/JBIES-24-00042

\bibitem{hassan2024}
Hassan N, Nayak A, Goh E, et al.
Addressing the challenge of governance, risk, and compliance in responsible AI.
\textit{IEEE Engineering Management Review}. 2024;52(3):90--100.
doi:10.1109/EMR.2024.3407478

\bibitem{croxford2025}
Croxford E, Gao M, Pellegrini E, et al.
Evaluating clinical AI summaries with large language models as judges.
\textit{npj Digital Medicine}. 2025;8:640.
doi:10.1038/s41746-025-02005-2

\bibitem{arora2025}
Arora R, Chaurasia A, Pfeffer MA, et al.
Introducing HealthBench.
OpenAI; 2025.
Available from: \url{https://openai.com/index/healthbench/}

\bibitem{tam2024}
Tam TYC, Sivarajkumar S, Kapoor S, et al.
A framework for human evaluation of large language models in healthcare derived from literature review.
\textit{npj Digital Medicine}. 2024;7:258.
doi:10.1038/s41746-024-01258-7

\bibitem{wang2025}
Wang W, et al.
An evaluation framework for ambient digital scribing tools in clinical applications.
\textit{npj Digital Medicine}. 2025;8:358.
doi:10.1038/s41746-025-01622-1

\bibitem{babic2025}
Babic B, et al.
A general framework for governing marketed AI/ML medical devices.
\textit{npj Digital Medicine}. 2025;8:328.
doi:10.1038/s41746-025-01717-9

\bibitem{hyperscribe_product}
Canvas Medical.
Hyperscribe. 2025.
Available from: \url{https://canvasmedical.com/extensions/hyperscribe}

\bibitem{hyperscribe_github}
Canvas Medical.
canvas-hyperscribe. GitHub; 2025.
Available from: \url{https://github.com/canvas-medical/canvas-hyperscribe}

\bibitem{canvas_sdk}
Canvas Medical.
Canvas SDK Commands.
Available from: \url{https://docs.canvasmedical.com/sdk/commands/}

\bibitem{paper_a}
Shah A, Hines A, Downs A, Bajet D, et al.
Case-Specific Rubrics for Clinical AI Evaluation: Methodology, Validation, and LLM-Clinician Agreement Across 823 Encounters.
[Companion paper--preprint pending]

\bibitem{physionet}
Shah A, et al.
Expert-Annotated Clinical Encounters: A Benchmark Dataset for Evaluating AI-Generated Medical Documentation.
PhysioNet. [To be added upon release]

\bibitem{dataset_usage}
Shah A, et al.
Expert-Annotated Clinical Encounters: Dataset Usage. GitHub; 2026.
Available from: \url{https://github.com/canvas-medical/dataset-usage}

\bibitem{hussein2026governance}
Hussein R, Schein O, Bates DW, et al.
Advancing healthcare AI governance through a comprehensive maturity model based on systematic review.
\textit{npj Digital Medicine}. 2026;9:236. doi:10.1038/s41746-026-02418-7

\bibitem{wells2025fairai}
Wells BJ, Walji MF, Sittig DF, et al.
A practical framework for appropriate implementation and review of artificial intelligence (FAIR-AI) in healthcare.
\textit{npj Digital Medicine}. 2025;8:514. doi:10.1038/s41746-025-01900-y

\bibitem{afshar2025pragmatic}
Afshar M, Dligach D, Churpek MM, et al.
A novel playbook for pragmatic trial operations to monitor and evaluate ambient artificial intelligence in clinical practice.
\textit{NEJM AI}. 2025;2(9). doi:10.1056/AIdbp2401267

\bibitem{nejmrct2025ambient}
Tierney AA, Sirek A, Ufere NN, et al.
Randomized clinical trial of two ambient AI scribes for clinical documentation.
\textit{NEJM AI}. 2025. doi:10.1056/AIoa2501000

\bibitem{asgari2025hallucination}
Asgari E, Fernandes M, Laranjo L, et al.
A framework to assess clinical safety and hallucination rates of LLMs for medical text summarisation.
\textit{npj Digital Medicine}. 2025;8:274. doi:10.1038/s41746-025-01670-7

\end{thebibliography}
\end{document}